\documentclass[conference]{IEEEtran}
\PassOptionsToPackage{sort}{natbib}
\IEEEoverridecommandlockouts
\usepackage{cite}
\usepackage{amsmath,amssymb,amsthm,amsfonts}
\usepackage{algorithmic}
\usepackage{graphicx}
\usepackage{textcomp}
\usepackage{xcolor}
\usepackage[numbers]{natbib}
\PassOptionsToPackage{hyphens}{url}\usepackage{hyperref}
\usepackage{amsfonts}       
\usepackage{nicefrac}       
\usepackage{microtype}      
\usepackage{wrapfig,lipsum,booktabs}
\usepackage{multicol}
\usepackage{enumitem}
\usepackage{graphicx} 
\usepackage{subcaption}
\usepackage{algorithm}
\usepackage{algorithmic}
\usepackage{bm}
\usepackage{xcolor}
\usepackage{mathdots}
\usepackage{float}
\usepackage{xspace}
\usepackage{array}
\usepackage{amsthm}
\usepackage{enumitem}
\def\BibTeX{{\rm B\kern-.05em{\sc i\kern-.025em b}\kern-.08em
    T\kern-.1667em\lower.7ex\hbox{E}\kern-.125emX}}

\theoremstyle{plain}
\newtheorem{theorem}{Theorem}

\newtheorem{assump}{Assumption}

\newtheorem{corollary}[theorem]{Corollary}
\theoremstyle{definition}
\newtheorem{definition}[theorem]{Definition}

\DeclareMathOperator*{\argmin}{argmin}
\def\bw{{\mathbf{w}}}
\newcommand{\eqdef}{\mathrel{\mathop:}=}

\newcommand{\fedprox}{\texttt{FedProx}\xspace}
\newcommand{\fedavg}{\texttt{FedAvg}\xspace}
\newcommand{\feddane}{\texttt{FedDANE}\xspace}
\newcommand{\dane}{\texttt{DANE}\xspace}
\newcommand{\scaffold}{\texttt{SCAFFOLD}\xspace}

\begin{document}

\title{{\feddane:} A Federated Newton-Type Method}

\author{Tian Li$^\dagger$~~~Anit Kumar Sahu$^\ddagger$~~~Manzil Zaheer$^*$~~~ Maziar Sanjabi$^\mathparagraph$~~~Ameet Talwalkar$^{\dagger\mathsection}$~~~Virginia Smith$^\dagger$\\
\small $^\dagger$Carnegie Mellon University~$^\ddagger$Bosch Center for AI~$^*$Google Research~$^\mathparagraph$University of Southern California~$\mathsection$Determined AI\\
\small $^\dagger$\{tianli, talwalkar, smithv\}@cmu.edu,
\small $^\ddagger$anit.sahu@gmail.com, $^*$manzilz@google.com, $^\mathparagraph$maziar.sanjabi@gmail.com}

\maketitle

\begin{abstract}
Federated learning aims to jointly learn statistical models over massively distributed remote devices. 
In this work, we propose \feddane, an optimization method that we adapt from \dane~\cite{shamir2014communication,AIDE_reddi_16}, a method for classical distributed optimization, to handle the practical constraints of federated learning. We provide convergence guarantees for this method when learning over both convex and non-convex functions. Despite encouraging theoretical results, we find that the method has underwhelming performance empirically. In particular, through empirical simulations on both synthetic and real-world datasets, \feddane consistently underperforms baselines of \fedavg~\cite{mcmahan2016FedAvg} and \fedprox~\cite{tian2018federated} in realistic federated settings. We identify low device participation and statistical device heterogeneity as two underlying causes of this underwhelming performance, and conclude by suggesting several directions of future work.
\end{abstract}

\section{Introduction}

Federated learning is a distributed learning paradigm that considers training statistical models in heterogeneous networks of remote devices~\cite{mcmahan2016FedAvg, survey}. Learning a model while keeping data localized can provide both computational and privacy benefits compared to transmitting raw data across the network.

To handle heterogeneity and high communication costs in federated networks, a popular approach for federated optimization methods involves allowing for local updating and low participation~\cite{survey}.
One method along these lines is \fedavg~\cite{mcmahan2016FedAvg}, which has demonstrated robust empirical performance  in non-convex settings. 
\fedavg assumes only a small subset of devices (e.g., 1\% out of thousands to millions) participate in training at each communication round. Each selected device then performs variable amounts of local work before sending model updates back to the server, which can enable a flexible trade-off between communication and computation.

Although \fedavg performs well empirically, it can diverge when the data is statistically heterogeneous (i.e., generated in a non-identically distributed manner across the network)~\cite{mcmahan2016FedAvg,tian2018federated}. A recent approach, \fedprox~\cite{tian2018federated}, has attempted to mitigate this issue by adding a proximal term to the subproblem on each device, which helps to improve the stability of the method.

In this work, we take a similar approach to \fedprox, and draw inspiration from \dane and variants~\cite{shamir2014communication, AIDE_reddi_16}, which are popular methods developed for the distributed data center setting. 
In particular,~\citet{AIDE_reddi_16} propose inexact-\dane, a variant of \dane that allows for local updating, which is beneficial when communication is a bottleneck. Compared with \fedavg, \dane and inexact-\dane use a different local subproblem which includes two additional terms---a gradient correction term and a proximal term. As data is statistically heterogeneous in federated networks, these terms can potentially improve convergence by forcing model updates to be closer to the current global model, making the method more stable and amenable to theoretical analysis. Including the gradient correction term also allows the update to take on the form of an approximate Newton-type method, which can lead to provably improved convergence for certain well-behaved objectives~\cite{shamir2014communication}.

Despite the merits of (inexact) \dane, the method has not been analyzed in settings with statistically heterogeneous data or low participation amongst the devices, which are critical challenges in realistic federated networks. Indeed, at each communication round, \dane requires every device to collectively evaluate the  gradient of the global function. This is prohibitive in federated networks as it requires the server to communicate with each device in a potentially massive network, and does not allow for the case of devices dropping out. A natural way to address this issue is to approximate the gradient via a subsample of the devices. Based on this idea, we propose \feddane, a variant of inexact \dane for federated learning.\footnote{We note that the gradient correction term in \feddane was explored briefly in prior work of~\fedprox~\cite{tian2018federated} (Appendix B), though this work is the first to theoretically analyze \feddane and provide a systematic evaluation of the method in federated settings.} Similar to inexact \dane, \feddane inexactly solves an approximate Newton-type subproblem, but only collects gradient updates from a subset of devices at each round. 

We provide convergence guarantees for \feddane for both convex and non-convex functions in low participation settings, and allow for the scenario that each device generates data from a possibly differing distribution. 
Despite encouraging theoretical results, our empirical evaluation indicates that while \feddane is more expensive as it needs two rounds of communication for one update, it consistently underperforms \fedavg and \fedprox due to the inexact estimation of the full gradient and the statistical heterogeneity in the network. 
Our study highlights the drawbacks of the gradient correction term in \feddane, and suggests the superiority of \fedprox which leverages just the proximal term to achieve improved performance for federated optimization. Our work also suggests several directions of future work in federated optimization.

\section{Related Work}\label{sec:related_work}

{\bf \dane and Other Communication-efficient Distributed Methods.} Methods that employ local updating (i.e., computing and applying a variable number of updates locally, rather than just evaluating the gradients once and sending them back for aggregation) are a popular approach for improving communication-efficiency in distributed optimization. By solving the local subproblems inexactly at each round, such schemes enable a  flexible trade-off between communication and computation. For example,  COCOA~\cite{COCOA_Smith_2016} is a communication-efficient primal-dual framework that leverages duality to decompose the global objective into subproblems that can be solved inexactly. Several primal methods~\citep[e.g.,][]{shamir2014communication,AIDE_reddi_16,elastic_SGD_zhang_LeCun_2015,cooperative_SGD_Joshi_18,local_SGD_stich_18,zhou2017convergence}, including \dane~\cite{shamir2014communication} and inexact \dane~\cite{AIDE_reddi_16}, have also been proposed, and have the added benefit of being applicable to non-convex objectives. While these methods make a seemingly small change over standard mini-batch methods, they enable drastically improved performance in practice, and have been shown to achieve orders-of-magnitude speedups over mini-batch methods in real-world data center environments. This is especially critical in communication-constrained environments such as federated settings. 

\vspace{.5em}

{\bf Heterogeneity-aware Federated Optimization.} An important distinction between federated optimization and classical distributed optimization is the presence of \textit{heterogeneity}, i.e., non-identically distributed data and heterogeneous systems across the network. \citet{fed_multitask_smith_2017} propose a primal-dual optimization method that learns separate but related models for each device through a multi-task learning framework. This setup naturally captures statistical heterogeneity, and also considers systems issues such as stragglers in the method and theory. However, such an approach is not generalizable to non-convex problems.
There are several recent works that provide theoretical analysis specifically for federated optimization. \fedprox~\cite{tian2018federated} characterizes the convergence behavior under a dissimilarity assumption of local functions, while accounting for the low participation of devices. Other works analyze different methods with non-identically distributed data, but under different (possibly) limiting assumptions, such as using SGD as a specific local solver~\cite{li2019fedavgon}, full device participation~\cite{wang2018adaptive,yu2018parallel}, convexity~\cite{li2019fedavgon,wang2018adaptive,karimireddy2019scaffold}, or uniformly-bounded gradients~\cite{yu2018parallel,li2019fedavgon}. For instance, {\scaffold~\cite{karimireddy2019scaffold} is a recent method for federated optimization related to \dane where it maintains a similar gradient correction term in the local subproblem. However, its convergence results are limited to strongly convex functions, and the method has yet to be explored empirically.}  Our convergence analysis of \feddane also accounts for low device participation and data heterogeneity, and covers both convex and non-convex functions (Section~\ref{sec:theory}).

\section{Methods}\label{sec:methods}
In this section, we propose \feddane, a heterogeneity-aware federated optimization method. Before introducing \feddane (Section~\ref{sec:methods:feddane}), we first formally define the optimization objective we consider in this paper (Section~\ref{sec:methods:setup}), and provide some background on \fedavg and \dane (Section~\ref{sec:methods:prelimi}).

\subsection{Problem Setup}\label{sec:methods:setup}
Federated learning {typically} aims to minimize the empirical risk over heterogeneous data distributed across multiple devices:
\begin{align}
\label{eq:obj_fedavg}
\min_w~f(\bw) = \sum_{k=1}^N p_k F_k(\bw)= \mathbb{E}_k [F_k(\bw)],
\end{align}
where $N$ is the number of devices, $p_k$ $\geq$ $0$, and $\sum_k p_k=1$. In general, the local objectives measure the local empirical risk over possibly differing data distributions $\mathcal{D}_k$, i.e., $F_k(w) \eqdef \mathbb{E}_{x_k \sim \mathcal{D}_k}{f_k(w;x_k)}$, with $n_k$ samples available at each device $k$. Hence, we can set $p_k=\frac{n_k}{n}$, where $n=\sum_k n_k$ is the total number of data points on all devices. 
In this work, we consider the typical centralized setup where $N$ devices are connected to one central server.

\subsection{Preliminaries: \fedavg and \dane}\label{sec:methods:prelimi}
In \fedavg~\cite{mcmahan2016FedAvg}, a subset of devices are sampled, and perform variable iterations of SGD to solve their local subproblems inexactly at each communication round. In particular, each selected device $k$ runs $E$ epochs of SGD on the local function $F_k$ to obtain local updates, then sends the updates back for aggregation in a synchronous manner. The details are summarized in Algorithm~\ref{alg:FEDAVG}. 

\begin{algorithm}[h]
    \begin{algorithmic}[1]
	\caption{Federated Averaging (\fedavg)}
	\label{alg:FEDAVG}
	\STATE {\bf Input:}  $K$, $T$, $\eta$, $E$, $\bw^0$, $N$, $p_k$, $k=1,\cdots, N$
	\FOR  {$t=1, \cdots, T$}
		\STATE Server selects a subset $S_t$ of $K$ devices at random (each device $k$ is chosen with probability $p_k$)
		\STATE Server sends $\bw^{t-1}$ to all chosen devices 
		\STATE Each device $k \in S_t$ updates $\bw^{t-1}$ for $E$ epochs of SGD on $F_k$ with step-size $\eta$ to obtain $\bw_k^{t}$
		\STATE Each device $k \in S_t$ sends $\bw_k^{t}$ back to the server
		\STATE Server aggregates the $\bw$'s as {$\bw^{t} = \frac{1}{K}\sum_{k \in S_t} \bw_k^{t}$}
	\ENDFOR
	\end{algorithmic}
\end{algorithm}

In data center settings, \dane~\cite{shamir2014communication} and its inexact variants~\cite{AIDE_reddi_16} are another set of approaches which have been analyzed in depth. In its simplest form, \dane has each worker $k$ solve the following subproblem:
\begin{align}\label{eq:dane}
    \bw_k^t = \argmin_{\bw} F_k(\bw) & +  \left\langle \nabla f(\bw^{t-1}) - \nabla F_k({\bw}^{t-1}),  \bw-\bw^{t-1} \right\rangle \nonumber \\ & + \frac{\mu}{2} \left\|\bw-\bw^{t-1}\right\|^2.
\end{align}
Similarly, after each worker solves its subproblem, the central server collects those updates and aggregates them to obtain $\bw^t$. The update is in fact a two-step process, as~\eqref{eq:dane} requires the workers to first collectively compute the overall gradient of the function, $\nabla f(\bw^{t-1})$, and can be interpreted as a distributed variant of SVRG~\cite{AIDE_reddi_16}. Inexact \dane allows the flexibility of solving~\eqref{eq:dane} inexactly~\cite{AIDE_reddi_16}. Based on inexact \dane, we next introduce \feddane.

\subsection{Proposed Method: \feddane}\label{sec:methods:feddane}

\begin{algorithm}[h]
\setlength{\abovedisplayskip}{0pt}
\setlength{\belowdisplayskip}{0pt}
\setlength{\abovedisplayshortskip}{0pt}
\setlength{\belowdisplayshortskip}{0pt}
    \begin{algorithmic}[1]
	\caption{Proposed method: \feddane}
	\label{alg:feddane}
	\STATE {\bf Input:}  $K$, $T$, $\eta$, $E$, $\bw^0$, $N$, $p_k$, $k=1,\cdots, N$
	\FOR  {$t=1, \cdots, T$}
		\STATE Server selects a subset $S_t$ of $K$ devices at random (each device $k$ is chosen with probability $p_k$)
		\STATE Server sends $\bw^{t-1}$ to all chosen devices 

		    \STATE Each selected device computes $\nabla F_k({\bw}^{t-1})$ and sends it to the central server
		    \STATE The server aggregates the gradients into
		    \begin{align*}
		        g_t = \frac{1}{K} \sum_{k \in S_t} \nabla F_k({\bw}^{t-1})
		    \end{align*}

		\STATE Server selects another subset $S_t'$ of $K$ devices at random; each device $k \in S_t'$ solves the following subproblem inexactly to obtain ${\bw}_k^{t}$:
		\begin{align*}
		    \bw_k^{t} = \argmin_{\bw} F_k(\bw) &+ \left\langle g_t - \nabla F_k(\bw^{t-1}), \bw-\bw^{t-1}\right\rangle \\ &+\frac{\mu}{2} \left\|\bw-\bw^{t-1}\right\|^2
		\end{align*}
		\STATE Each device $k \in S_t'$ sends $\bw_k^{t}$ back to the server
		\STATE Server aggregates the $\bw$'s as {$\bw^{t} = \frac{1}{K}\sum_{k \in S_t} \bw_k^{t}$}
	\ENDFOR
	\end{algorithmic}
\end{algorithm}

The inexact \dane method mentioned above cannot be directly applied to federated settings. One critical challenge is that computing the full gradient $\nabla f(\bw^{t-1})$ requires the server to communicate with all the devices and then average the local gradients, which is infeasible in massive federated networks. 

In \feddane, we propose to approximate the full gradients using a subset of gradients from randomly sampled devices. Collecting the gradients from a subset $S_t$ ($|S_t|=K$) of devices at each iteration $t$ yields: 
\begin{align*}
    \nabla f(\bw^{t-1}) \approx g_t = \frac{1}{K} \sum_{k \in S_t} \nabla F_k(\bw^{t-1}).
\end{align*}
After computing $g_t$, \feddane selects another subset of devices where each device $k \in S_t$ solves the following subproblem inexactly:
\begin{align}\label{eq:obj}
    \bw_k^t = \argmin_{\bw} F_k(\bw) &+ \left\langle g_t - \nabla F_k({\bw}^{t-1}), \bw-\bw^{t-1} \right\rangle \nonumber\\
     &+ \frac{\mu}{2} \left\|\bw-\bw^{t-1}\right\|^2.
\end{align}
The server then aggregates the updates from the selected devices. See Algorithm~\ref{alg:feddane} for details. We note that one limitation of \feddane is that each outer iteration incurs two rounds of communication, making it less efficient than \fedavg and \fedprox. This leads us to suggest a variant of \feddane leveraging a pipelined approach to perform one update in a single round of communication (see Section~\ref{sec:exp:discussion} for more discussions). However, as we will see in our empirical valuation (Section~\ref{sec:experiments}), even the  less efficient (and more accurate) \feddane proposed here results in inferior practical performance compared to \fedavg and \fedprox.

\section{Analysis}\label{sec:theory}

We now provide our convergence analysis of \feddane for both convex and non-convex problems.
Recall that \feddane allows each selected device to solve a subproblem inexactly at each updating round to reduce communication. We first formally define a parameter $\gamma$ to quantify the inexactness, which will be used throughout our analysis. 
\begin{definition}[$\gamma$-inexact Solution]\label{def:inexact}
We say that ${\bw}^t$ is a $\gamma$-inexact minimizer of~\eqref{eq:obj} if $\left\|\bw^t - \underline{\bw}^t\right\| \leq \gamma \left\|\underline{\bw}^t-\bw^{t-1}\right\|$, where $\gamma \in [0,1)$, and $\underline{\bw}^t$ is the exact minimizer of~\eqref{eq:obj}. Note that a smaller $\gamma$ corresponds to higher accuracy.
\end{definition}

In order to quantify the dissimilarity between devices in a federated network, following~\citet{tian2018federated}, we define $B$-local dissimilarity as follows.

\begin{definition}[$B$-local Dissimilarity]\label{def: similarity}
The local functions $F_k$ are $B$-locally dissimilar at $\bw$ if ${\mathbb{E}_k{\|\nabla F_k(\bw)\|^2}}\! \leq\!{\|\nabla f(\bw)\|^2}B^2$. We further define $B(\bw)\!=\! \sqrt{\frac{\mathbb{E}_{k}{\|\nabla F_k(\bw)\|^2}}{\|\nabla f(\bw)\|^2}}$ for $\|\nabla f(\bw)\|\!\neq\!0$. 
\end{definition}
When the devices are homogeneous with I.I.D. data, $B(\bw)=1$ for every $\bw$. The more heterogeneous the data are in the network, the larger $B(\bw)$ is. As discussed later, our convergence results are a function of the device dissimilarity bound $B$.

\subsection{Convex Case}
We first investigate the convergence results for  convex $F_{k}$'s.
\begin{theorem}[Sufficient Decrease]
	\label{th:convex}
	 Assume $F_k$'s are convex, and have $L$-Lipschitz continuous gradients. Moreover, assume $B$-dissimilarity is bounded by $B$ at point $\bw^{t-1}$. Given the inexact criterion in Definition~\ref{def:inexact}, if $\mu, \gamma, L,$ and $B$ satisfy
	\begin{align*}
    \rho = &\bigg(\frac{2-3\gamma}{2\mu} -\frac{2L(1+\gamma)^2+3L}{2\mu^2} \\ & - (B^2-1) \left(\frac{L(1+\gamma)^2+L}{\mu^2} + \frac{\gamma}{\mu}\right)\bigg) >0,
    \end{align*}	
    then at iteration $t$ of Algorithm~\ref{alg:feddane}, we have the following expected decrease in the global objective:
	\begin{align*}
	\mathbb{E}_{S_{t}}\left[f(\mathbf{\mathbf{w}}^{t}) \right]\le f(\mathbf{\mathbf{w}}^{t-1})-\rho \left\|\nabla f(\bw^{t-1})\right\|^2,
	\end{align*}
	where $S_{t}$ represents the distribution of a set of random devices selected at time $t$.
\end{theorem}

We defer the readers to Appendix~\ref{appen:theorem3} for a complete proof. At a high-level, we first use the $\gamma$-inexactness and other assumptions to attain a decrease in the objective, then take an expectation over randomly selected devices and apply the bounded $B$-dissimilarity to obtain the above results.

\begin{corollary}[Convergence: Convex Case] Let the assertions of Theorem~\ref{th:convex} hold. In addition, let $\gamma$ = 0, i.e., all the local problems are solved exactly, if $1 \ll B$ then we choose $\mu \approx 5LB^2$ from which it follows that $\rho \approx \frac{3}{25LB^2}$.

\end{corollary}

\subsection{Non-convex Case}
We have the following convergence characterization for non-convex functions.
	
\begin{theorem}[Sufficient Decrease]
	\label{th:non-convex}
	Assume $F_k$'s are non-convex, and have $L$-Lipschitz continuous gradients. Moreover, assume there exists a $\lambda$ such that $\lambda \mathbf{I} + \nabla ^2 F_k(w) \succ 0$, with $\mu-\lambda>0$. Assume $B$-dissimilarity is bounded by $B$ at point $\bw^{t-1}$. Given the inexact criterion in Definition~\ref{def:inexact}, if $\mu, \gamma, L,$ and $B$ satisfy
	\begin{align*}
	 \rho &= \bigg(\frac{1}{\mu}-\frac{3\gamma}{2(\mu-\lambda)} - \frac{L(1+\gamma)^2}{(\mu-\lambda)^2} - \frac{3L}{2\mu(\mu-\lambda)} \\ &- \left(B^2-1\right) \left(\frac{L(1+\gamma)^2}{(\mu-\lambda)^2}+\frac{L}{\mu(\mu-\lambda)}+\frac{\gamma}{\mu-\lambda}\right)\bigg) > 0
	\end{align*}
	then at iteration $t$ of Algorithm~\ref{alg:feddane}, we have the following expected decrease in the global objective:
	\begin{align*}
	    \mathbb{E}_{S_t}[f(\bw^t)] \leq f(\bw^{t-1})-\rho \|\nabla f(\bw^{t-1})\|^2,
	\end{align*}
	where $S_{t}$ represents the devices randomly selected at time $t$.
\end{theorem}

The proof (Appendix~\ref{appen:theorem5}) is similar to the proof for Theorem~\ref{th:convex}. Now we can use the above sufficient decrease to the characterize the rate of convergence to the set of
approximate stationary solutions $\{w~|~ \mathbb{E}\left[\|\nabla f(\bw^t)\|^2\right] \leq \epsilon\}$.

\begin{theorem}[Convergence: Non-convex Case] Let the assumption Theorem~\ref{th:non-convex} hold at each iteration of \feddane. Moreover, $f(\bw^0)-f^{*}=\Delta$. Then, after $T=O(\frac{\Delta}{\rho \epsilon})$ iterations, we have $\frac{1}{T}\sum_{t=1}^T \mathbb{E}\left[\|\nabla f(\bw^t)\|^2\right] \leq \epsilon$.
\end{theorem}

Note that the convergence rates of \feddane derived here recover the results in \fedprox~\cite{tian2018federated}, which are also asymptotically the same  as SGD~\cite{ghadimi2013stochastic}.

\subsection{Device-specific Constants}

While the previous results assume the same constants $L$ (the Lipschitz constant of gradients), $\mu$ (the penalty constant of the proximal term), and $\gamma$ (the degree of inexactness) across all devices, we can easily extend the analysis to allow for variable constants across devices. 
	
\begin{theorem}[Convergence with Device-specific Constants]\label{th:device_specific}
	Assume $F_k$'s are convex, and have $L_k$-Lipschitz continuous gradients. Moreover, assume $B$-dissimilarity is bounded by $B$ at point $\bw^{t-1}$. Given the inexact criterion in Definition~\ref{def:inexact}, if  constants $\mu_k, \gamma_k, L_k,$ and $B$ are chosen such that
	\begin{align*}
    \rho &=\bigg(\frac{1}{K_t}\sum_{k=1}^{K_t} \left(\frac{1}{\mu_k}- \frac{3\gamma_k}{2\mu_k}-\frac{L_k\left(1+\gamma_k\right)^2}{\mu_k^2}-\frac{3L_k}{2\mu_k^2}\right) \\ & -\frac{1}{K_t}\sum_{k=1}^{K_t} \left(\frac{L\left(1+\gamma_k\right)^2}{\mu_k^2} + \frac{L_k}{\mu_k^2}+\frac{\gamma_k}{\mu_k}\right)\left(B^2-1\right)\bigg)>0,
    \end{align*}	
    then at iteration $t$ of Algorithm~\ref{alg:feddane}, we have the following expected decrease in the global objective:
	\begin{align*}
        \mathbb{E}_{S_t}[f(\bw^t)] \leq f(\bw^{t-1})-\rho \|\nabla f(\bw^{t-1})\|^2,
	\end{align*}
	where $S_{t}$ represents the distribution of a set of random devices selected at time $t$.
\end{theorem}

See Appendix~\ref{appen:theorem7} for a full proof. Note that our analysis is general in that it is agnostic of any specific local solver, and covers both cases of sampling devices  with and without replacement.

\section{Experiments}\label{sec:experiments}

\begin{figure*}[t]
    \centering
    \includegraphics[width=1.0\textwidth]{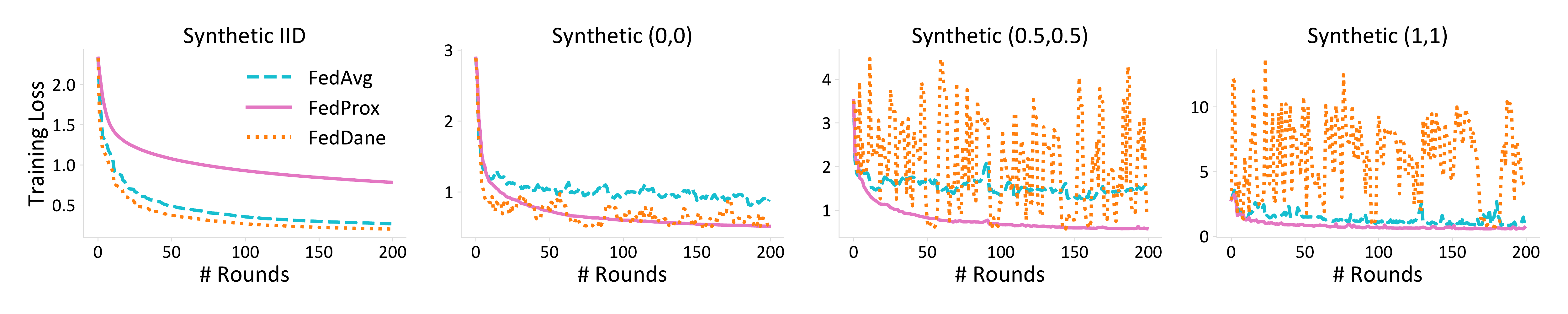}
     \includegraphics[width=0.32\textwidth]{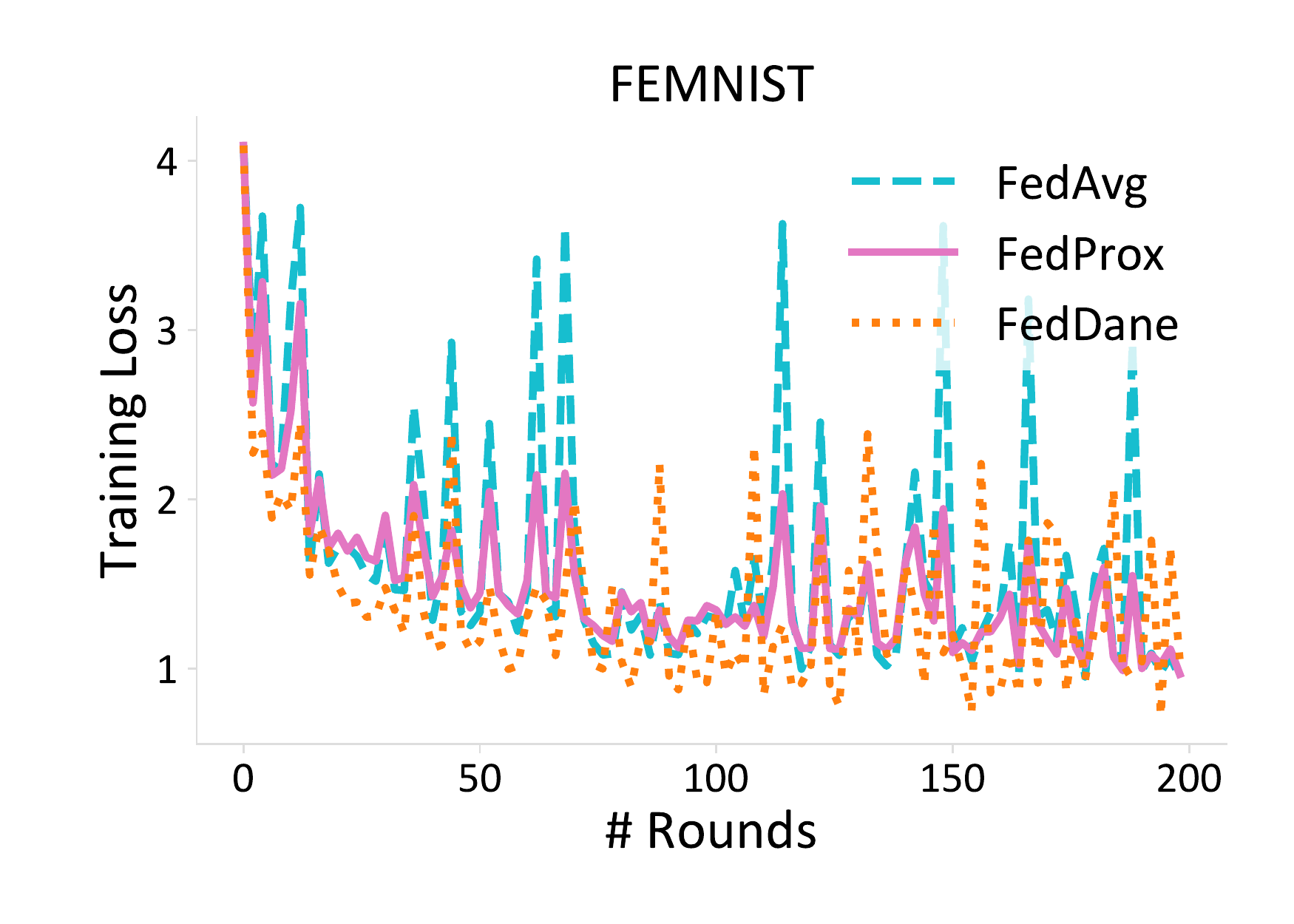}
    \includegraphics[width=0.34\textwidth]{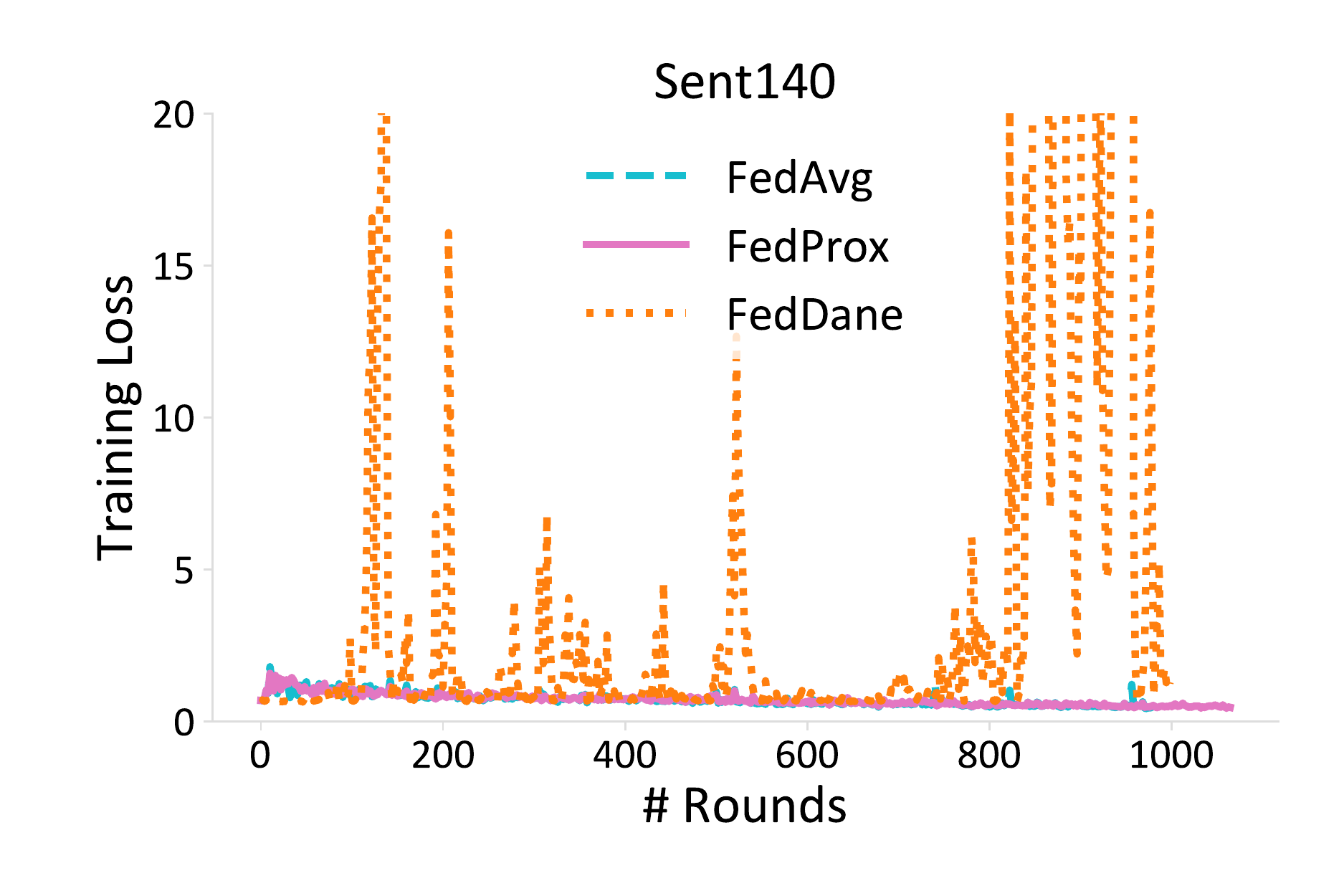}
    \includegraphics[width=0.32\textwidth]{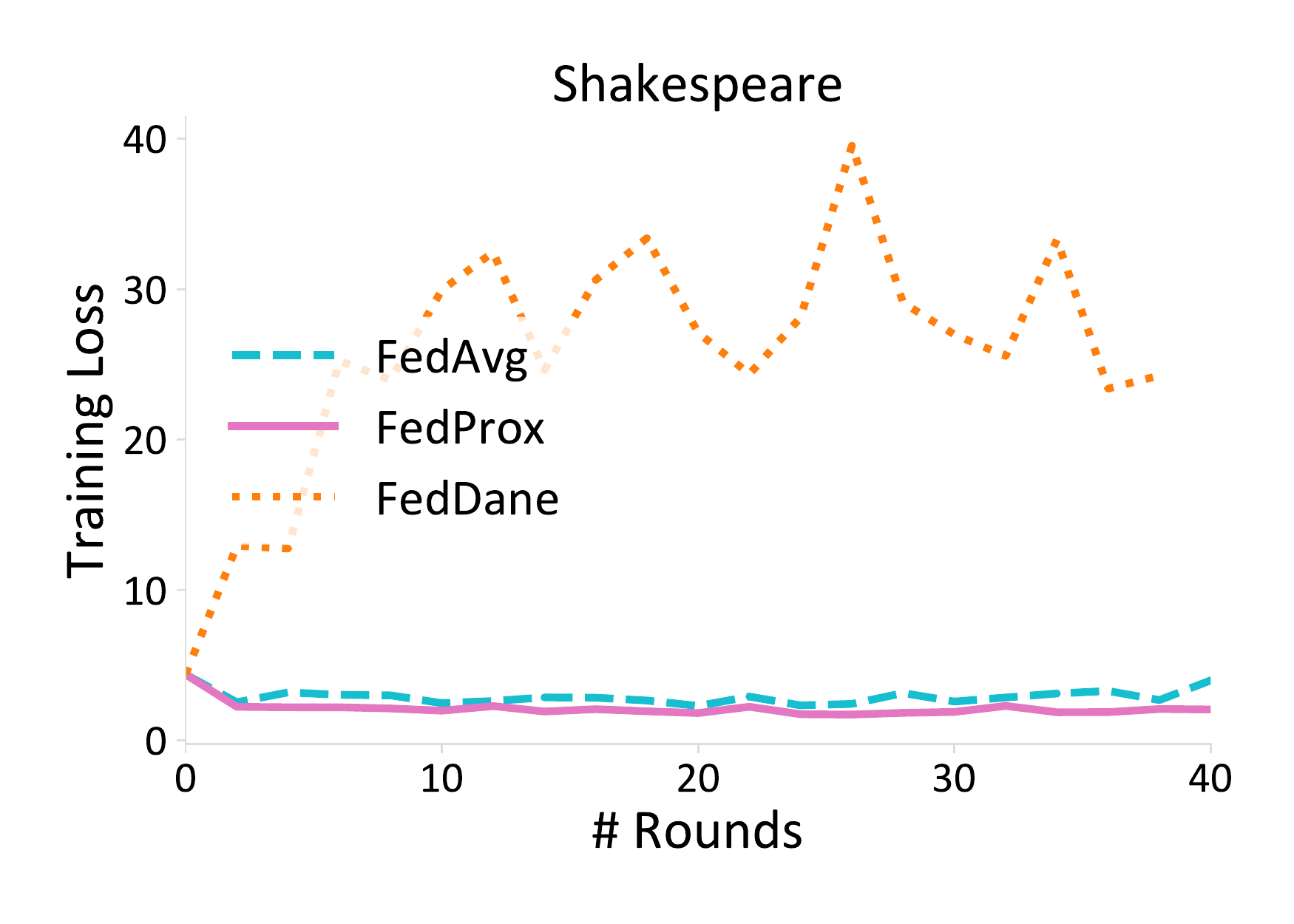}
    \caption{Convergence of \feddane compared with \fedavg and \fedprox. For synthetic datasets in the first row, from left to right, data are becoming more heterogeneous. Except for the perfect I.I.D. dataset (Synthetic IID), \feddane underperforms both \fedavg and \fedprox on all datasets---either converging more slowly or diverging.}
    \label{fig:results}
\end{figure*}

\subsection{Experimental Setup} 
{\bf Datasets.} We evaluate the performance of \feddane using both synthetic and real-world federated datasets. The datasets are curated from the LEAF benchmark~\cite{caldas2018leaf} as well as previous work on federated learning~\cite{tian2018federated}. In particular, we use a set of synthetic datasets with varying degrees of data heterogeneity following the setup in~\citet{tian2018federated}. We also study three real datasets in LEAF: FEMNIST for image classification with a convex model, Shakespeare for next-character prediction, and Sent140 for sentiment analysis, both with non-convex deep neural network models. These datasets are naturally partitioned into different devices in the network~\cite{caldas2018leaf}. Data statistics are summarized in Table~\ref{table:data} below.

\begin{table}[h]
    \caption{Statistics of three real federated datasets.}
	\begin{center}
		\label{table:data}
		\begin{tabular}{ lllll } 
			\toprule
			\textbf{Datasets} & \textbf{\# Devices} & \textbf{\# Samples} & 
			\multicolumn{2}{l}{\textbf{\# Samples/device}} \\
			\cmidrule(l){4-5}
			& &  & mean & stdev \\
			\hline
			FEMNIST  & 200 & 18,345 & 92 & 159 \\
			Sent140 & 772 &  40,783 & 53 & 32 \\
			Shakespeare & 143 & 517,106 & 3,616 & 6,808 \\
			\bottomrule
		\end{tabular}
	\end{center}
\end{table}

{\bf Implementation \& Hyper-parameters.} We implement all code in Tensorflow~\citep{abadi2016tensorflow}, simulating a federated setup where $N$ devices ($N$ is the total number of devices shown in Table~\ref{table:data}) are connected with a central server. For \fedavg and \fedprox, we directly take the tuned hyper-parameters reported in~\cite{tian2018federated}. For \feddane, we use the same learning rates and batch sizes as in \fedavg on the same dataset. We tune $\mu$ (the penalty constant in the proximal term) for \feddane from a candidate set \{0, 0.001, 0.01, 0.1, 1\} and pick a best $\mu$ based on the training loss. All code,
data, and experiments are publicly available at \href{https://github.com/litian96/FedDANE}{\texttt{github.com/litian96/FedDANE}}.

\subsection{Evaluation Results}

We compare the convergence of \feddane with \fedavg and \fedprox. For each method, we select 10 devices at each updating round, and let each device perform $E$ epochs of local updates ($E=20$). We plot the training loss versus the updating rounds (treating two communication rounds in \feddane as one). The results are shown in Figure~\ref{fig:results}. We see that \feddane consistently performs worse than both \fedavg and \fedprox. This indicates that  statistical heterogeneity and low device participation (the inaccurate approximation of the full gradients)  may hurt the convergence of \feddane. We further investigate the effects of varying participating devices and show that whether selecting more devices to get a better approximation of the full gradients can lead to improved performance depends on the degree of data heterogeneity. We then create an extreme `unrealistic' setting that favors \feddane, where we select a large subset of devices (78\% of the total devices on average) and let each device perform only one epoch of local updates, trying to prevent local models from deviating too much from the global model.  Even in this unrealistic setting, the performance of \feddane is still disappointing. The results of all the additional experiments are provided in Appendix~\ref{appen:exp}. 

\subsection{Discussions}\label{sec:exp:discussion}

Despite encouraging theoretical results, \feddane demonstrates underwhelming empirical performance. This indicates that several of our theoretical assumptions may not hold in practical scenarios. These violations may include (1) the lowest eigenvalue of the Hessian $\nabla^2F_k(w)$ is too small, (2) the choice of $\mu$ does not make the local subproblem strongly convex, and (3) the choices of the constants $\mu, \gamma, L$ and $B$ may not guarantee sufficient decrease. More generally, the discrepancy between theory and practice suggests that the practical issues of low device participation and statistical heterogeneity in distributed optimization require careful theoretical consideration---for \feddane as well as for methods such as \fedavg and \fedprox. Developing a better understanding of this setting may help to enable improved empirical performance for the increasingly prevalent problem of federated learning.

We note that there are other possible variants of \dane that may address the drawbacks of \feddane. For instance, in order to mitigate the negative effects of the gradient correction term, we can consider decaying this term over the optimization process. The `decayed' \feddane will eventually reduce to \fedprox as the gradient correction term becomes closer to zero. Another limitation with the proposed \feddane method is that it requires two rounds of communication for one update. 
One could imagine a `pipelined' variant of \feddane where the overall gradient and the local model updates are transmitted together to the server. In this way, however, the selected devices have to use the stale gradients for the gradient correction term in the local subproblem. 
Exploring such variants is an interesting direction of future research.

\section*{Acknowledgment}

This work was supported in part by DARPA FA875017C0141, the National Science Foundation
grants IIS1705121 and IIS1838017, an Okawa Grant, a Google Faculty Award, an Amazon Web
Services Award,  a JP Morgan A.I. Research Faculty Award, a Carnegie Bosch Institute Research Award, and the CONIX Research Center, one of
six centers in JUMP, a Semiconductor Research Corporation (SRC) program sponsored by DARPA.
Any opinions, findings, and conclusions or recommendations expressed in this material are those of
the author(s) and do not necessarily reflect the views of DARPA, the National Science Foundation, or
any other funding agency.

{
\bibliographystyle{abbrvnat}
\bibliography{ref}}

\newpage
\onecolumn
\section*{Appendix}

\subsection{Proof for Theorem~\ref{th:convex}} \label{appen:theorem3}
\begin{proof}
	We have by the Lipschitz continuity of the gradients:
	{\small\begin{align}\label{eq:th1_pr_1}
	&f(\mathbf{\mathbf{w}}^{t}) \leq f(\mathbf{\mathbf{w}}^{t-1}) + \langle \nabla f(\mathbf{\mathbf{w}}^{t-1}), \mathbf{\mathbf{w}}^{t}-\mathbf{\mathbf{w}}^{t-1}\rangle + \frac{L}{2}\left\|\mathbf{\mathbf{w}}^{t}-\mathbf{\mathbf{w}}^{t-1}\right\|^{2}\nonumber\\
	&\leq f(\mathbf{\mathbf{w}}^{t-1}) + \langle \nabla f(\mathbf{\mathbf{w}}^{t-1}), \underline{\mathbf{w}}^{t}-\mathbf{\mathbf{w}}^{t-1}\rangle+ \langle \nabla f(\mathbf{\mathbf{w}}^{t-1}), \mathbf{\mathbf{w}}^{t}-\underline{\mathbf{w}}^{t}\rangle + \frac{L}{2}\left\|\mathbf{\mathbf{w}}^{t}-\mathbf{\mathbf{w}}^{t-1}\right\|^{2}
	\end{align}}
	
	By optimality conditions, we have that $\underline{\mathbf{w}}^{t}$ satisfies
	{\small\begin{align}
	\label{eq:th1_pr_2}
	\nabla F_{k}\left(\underline{\mathbf{w}}^{t}\right)+\mathbf{g}_{t} - \nabla F_{k}\left(\mathbf{\mathbf{w}}^{t-1}\right) + \mu\left(\underline{\mathbf{w}}^{t}-\mathbf{\mathbf{w}}^{t-1}\right) = 0.
	\end{align}}
	
	We denote the local subproblem \eqref{eq:obj} as $P_t(w)$.
	We also note that, $P_t(\bw)$ is $\mu$-strongly convex,
	{\small\begin{align}
	\label{eq:th1_pr_3}
	\mu\left\|\underline{\mathbf{w}}^{t}-\mathbf{\mathbf{w}}^{t-1}\right\| \le \left\|\nabla P_t(\mathbf{\mathbf{w}}^{t-1}) \right\| = \left\|\mathbf{g}_{t}\right\|.
	\end{align}}
	
	We derive a bound for $\left\|\mathbf{\mathbf{w}}^{t}-\mathbf{\mathbf{w}}^{t-1}\right\|$ next.
	{\small\begin{align}
	\label{eq:th1_pr_3.5}
	\left\|\mathbf{\mathbf{w}}^{t}-\mathbf{\mathbf{w}}^{t-1}\right\| \le \left\|\underline{\mathbf{w}}^{t}-\mathbf{\mathbf{w}}^{t-1}\right\|+\left\|\underline{\mathbf{w}}^{t}-\mathbf{\mathbf{w}}^{t}\right\| \le (1+\gamma)\left\|\underline{\mathbf{w}}^{t}-\mathbf{\mathbf{w}}^{t-1}\right\|.
	\end{align}}
	
	Using \eqref{eq:th1_pr_2}-\eqref{eq:th1_pr_3} in \eqref{eq:th1_pr_1}, we have,
	{\small\begin{align}
	\label{eq:th1_pr_4}
	&f(\mathbf{\mathbf{w}}^{t}) \leq f(\mathbf{\mathbf{w}}^{t-1}) + \langle \nabla f(\mathbf{\mathbf{w}}^{t-1}), \underline{\mathbf{w}}^{t}-\mathbf{\mathbf{w}}^{t-1}\rangle+ \langle \nabla f(\mathbf{\mathbf{w}}^{t-1}), \mathbf{\mathbf{w}}^{t}-\underline{\mathbf{w}}^{t}\rangle + \frac{L}{2}\left\|\mathbf{\mathbf{w}}^{t}-\mathbf{\mathbf{w}}^{t-1}\right\|^{2}\nonumber\\
	&\leq f(\mathbf{\mathbf{w}}^{t-1}) - \frac{1}{\mu} \langle \nabla f(\mathbf{\mathbf{w}}^{t-1}), \nabla F_{k}\left(\underline{\mathbf{w}}^{t}\right)+\mathbf{g}_{t} - \nabla F_{k}\left(\mathbf{\mathbf{w}}^{t-1}\right)\rangle + \left\|\nabla f(\mathbf{\mathbf{w}}^{t-1})\right\|\left\|\mathbf{\mathbf{w}}^{t}-\underline{\mathbf{w}}^{t}\right\| + \frac{L(1+\gamma)^{2}}{2\mu^{2}}\left\|\mathbf{g}_{t}\right\|^{2}\nonumber\\
	&\leq f(\mathbf{\mathbf{w}}^{t-1})-\frac{\nabla^{\top} f(\mathbf{\mathbf{w}}^{t-1})\mathbf{g}_{t}}{\mu}+\frac{L}{\mu}\left\|\nabla f(\mathbf{\mathbf{w}}^{t-1})\right\|\left\|\underline{\mathbf{w}}^{t}-\mathbf{\mathbf{w}}^{t-1}\right\|+\gamma\left\|\nabla f(\mathbf{\mathbf{w}}^{t-1})\right\|\left\|\underline{\mathbf{w}}^{t}-\mathbf{\mathbf{w}}^{t-1}\right\|+ \frac{L(1+\gamma)^{2}}{2\mu^{2}}\left\|\mathbf{g}_{t}\right\|^{2}\nonumber\\
	&\leq f(\mathbf{\mathbf{w}}^{t-1})-\frac{\nabla^{\top} f(\mathbf{\mathbf{w}}^{t-1})\mathbf{g}_{t}}{\mu}+\frac{L}{2\mu^{2}}\left(\left\|\nabla f(\mathbf{\mathbf{w}}^{t-1})\right\|^{2}+\left\|\mathbf{g}_{t}\right\|^{2}\right) +\frac{\gamma}{2\mu}\left(\left\|\nabla f(\mathbf{\mathbf{w}}^{t-1})\right\|^{2}+\left\|\mathbf{g}_{t}\right\|^{2}\right)+ \frac{L(1+\gamma)^{2}}{2\mu^{2}}\left\|\mathbf{g}_{t}\right\|^{2}.
\end{align}}

Taking expectation with respect to the randomly chosen devices $S_t$ yields
{\small\begin{align}\label{eq:13}
	 \mathbb{E}_{S_{t}}\left[f(\mathbf{\mathbf{w}}^{t}) \right] \le  & f(\mathbf{\mathbf{w}}^{t-1})-\left(1-\frac{3\gamma}{2}\right)\frac{\left\|\nabla f(\mathbf{\mathbf{w}}^{t-1})\right\|^{2}}{\mu}+
	 \left(\frac{2L(1+\gamma)^{2}}{2\mu^{2}}+\frac{3L}{2\mu^{2}}\right)\left\|\nabla f(\mathbf{\mathbf{w}}^{t-1})\right\|^{2}\nonumber\\&+\left(\frac{L(1+\gamma)^{2}}{\mu^{2}}+\frac{L}{\mu^{2}}+\frac{\gamma}{\mu}\right)\mathbb{E}_{S_{t}}\left[\left\|\mathbf{g}_{t}-\nabla f(\mathbf{\mathbf{w}}^{t-1})\right\|^{2}\right],
\end{align}}

	where in the last step, we used the inequality that
	{\small\begin{align}
	\label{eq:ineq}
	\left\|\mathbf{g}_{t}\right\|^{2} = \left\|\mathbf{g}_{t}-\nabla f(\mathbf{\mathbf{w}}^{t-1})+\nabla f(\mathbf{\mathbf{w}}^{t-1})\right\|^{2} \le 2\left\|\nabla f(\bw^{t-1})\right\|^{2}+2\left\|\mathbf{g}_{t}-\nabla f(\mathbf{\mathbf{w}}^{t-1})\right\|^{2}.
	\end{align}}

Note that 
{\small\begin{align}
    \mathbb{E}_{S_t} \left[\left\|\mathbf{g}_t - \nabla f(\bw^{t-1})\right\|^2\right] = \mathbb{E}_{S_t}\left[\|\mathbf{g}_t\|^2\right] - \left\|\nabla f(\bw^{t-1})\right\|^2 \leq \mathbb{E}_k\left[ \| \nabla F_k(\bw^{t-1}) \|^2\right] - \left\|\nabla f(\bw^{t-1})\right\|^2 \leq (B^2-1) \left\|f(\bw^{t-1})\right\|^2.
\end{align}}

Plugging into \eqref{eq:13}, we get
{\small\begin{align}
    \mathbb{E}_{S_t}\left[f(\bw^t)\right] \leq f(\bw^{t-1}) - \rho \left\|\nabla f(\bw^{t-1})\right\|^2,
\end{align}}

where
{\small\begin{align}
    \rho = \frac{2-3\gamma}{2\mu} - \frac{2L(1+\gamma)^2+3L}{2\mu^2} - \left(B^2-1\right) \left(\frac{L\left(1+\gamma\right)^2+L}{\mu^2} + \frac{\gamma}{\mu}\right).
\end{align}}

\end{proof}

\subsection{Proof for Theorem~\ref{th:non-convex}}\label{appen:theorem5}

\begin{proof}
	We have by the Lipschitz continuity of the gradients:
	{\small\begin{align}
	\label{eq:th2_pr_1}
	&f(\mathbf{\mathbf{w}}^{t}) \leq f(\mathbf{\mathbf{w}}^{t-1}) + \langle \nabla f(\mathbf{\mathbf{w}}^{t-1}), \mathbf{\mathbf{w}}^{t}-\mathbf{\mathbf{w}}^{t-1}\rangle + \frac{L}{2}\left\|\mathbf{\mathbf{w}}^{t}-\mathbf{\mathbf{w}}^{t-1}\right\|^{2}\nonumber\\
	&\leq f(\mathbf{\mathbf{w}}^{t-1}) + \langle \nabla f(\mathbf{\mathbf{w}}^{t-1}), \underline{\mathbf{w}}^{t}-\mathbf{\mathbf{w}}^{t-1}\rangle+ \langle \nabla f(\mathbf{\mathbf{w}}^{t-1}), \mathbf{\mathbf{w}}^{t}-\underline{\mathbf{w}}^{t}\rangle + \frac{L}{2}\left\|\mathbf{\mathbf{w}}^{t}-\mathbf{\mathbf{w}}^{t-1}\right\|^{2}
	\end{align}}
	
	By optimality conditions, we have that $\underline{\mathbf{w}}^{t}$ satisfies
	{\small\begin{align}
	\label{eq:th2_pr_2}
	\nabla F_{k}\left(\underline{\mathbf{w}}^{t}\right)+\mathbf{g}_{t} - \nabla F_{k}\left(\mathbf{\mathbf{w}}^{t-1}\right) + \mu\left(\underline{\mathbf{w}}^{t}-\mathbf{\mathbf{w}}^{t-1}\right) = 0.
	\end{align}}

	We denote the local subproblem \eqref{eq:obj} as $P_t(w)$.
	We also note that, $P_t(\bw)$ is $(\mu-\lambda)$-strongly convex, 
	{\small\begin{align}
	\label{eq:th2_pr_3}
	(\mu-\lambda)\left\|\underline{\mathbf{w}}^{t}-\mathbf{\mathbf{w}}^{t-1}\right\| \le \left\|\nabla P_t(\mathbf{\mathbf{w}}^{t-1}) \right\| = \left\|\mathbf{g}_{t}\right\|.
	\end{align}}
	
	Using \eqref{eq:th2_pr_2}-\eqref{eq:th2_pr_3} in \eqref{eq:th2_pr_1}, we have,
	{\small\begin{align}
	\label{eq:th2_pr_4}
	&f(\mathbf{\mathbf{w}}^{t}) \leq f(\mathbf{\mathbf{w}}^{t-1}) + \langle \nabla f(\mathbf{\mathbf{w}}^{t-1}), \underline{\mathbf{w}}^{t}-\mathbf{\mathbf{w}}^{t-1}\rangle+ \langle \nabla f(\mathbf{\mathbf{w}}^{t-1}), \mathbf{\mathbf{w}}^{t}-\underline{\mathbf{w}}^{t}\rangle + \frac{L}{2}\left\|\mathbf{\mathbf{w}}^{t}-\mathbf{\mathbf{w}}^{t-1}\right\|^{2}\nonumber\\
	&\leq f(\mathbf{\mathbf{w}}^{t-1}) - \frac{1}{\mu} \langle \nabla f(\mathbf{\mathbf{w}}^{t-1}), \nabla F_{k}\left(\underline{\mathbf{w}}^{t}\right)+\mathbf{g}_{t} - \nabla F_{k}\left(\mathbf{\mathbf{w}}^{t-1}\right)\rangle + \left\|\nabla f(\mathbf{\mathbf{w}}^{t-1})\right\|\left\|\mathbf{\mathbf{w}}^{t}-\underline{\mathbf{w}}^{t}\right\| + \frac{L(1+\gamma)^{2}}{2(\mu-\lambda)^{2}}\left\|\mathbf{g}_{t}\right\|^{2}\nonumber\\
	&\leq f(\mathbf{\mathbf{w}}^{t-1})-\frac{\nabla^{\top} f(\mathbf{\mathbf{w}}^{t-1})\mathbf{g}_{t}}{\mu}+\frac{L}{\mu}\left\|\nabla f(\mathbf{\mathbf{w}}^{t-1})\right\|\left\|\underline{\mathbf{w}}^{t}-\mathbf{\mathbf{w}}^{t-1}\right\|+\gamma\left\|\nabla f(\mathbf{\mathbf{w}}^{t-1})\right\|\left\|\underline{\mathbf{w}}^{t}-\mathbf{\mathbf{w}}^{t-1}\right\|+ \frac{L(1+\gamma)^{2}}{2(\mu-\lambda)^{2}}\left\|\mathbf{g}_{t}\right\|^{2}\nonumber\\
	&\leq f(\mathbf{\mathbf{w}}^{t-1})-\frac{\nabla^{\top} f(\mathbf{\mathbf{w}}^{t-1})\mathbf{g}_{t}}{\mu}+\frac{L}{2\mu(\mu-\lambda)}\left(\left\|\nabla f(\mathbf{\mathbf{w}}^{t-1})\right\|^{2}+\left\|\mathbf{g}_{t}\right\|^{2}\right)\nonumber\\
	&+\frac{\gamma}{2(\mu-\lambda)}\left(\left\|\nabla f(\mathbf{\mathbf{w}}^{t-1})\right\|^{2}+\left\|\mathbf{g}_{t}\right\|^{2}\right)+ \frac{L(1+\gamma)^{2}}{2(\mu-\lambda)^{2}}\left\|\mathbf{g}_{t}\right\|^{2}\nonumber\\
	&\Rightarrow \mathbb{E}_{S_{t}}\left[f(\mathbf{\mathbf{w}}^{t}) \right]\le f(\mathbf{\mathbf{w}}^{t-1})-\left(1-\frac{3\gamma\mu}{2(\mu-\lambda)}\right)\frac{\left\|\nabla f(\mathbf{\mathbf{w}}^{t-1})\right\|^{2}}{\mu}+\left(\frac{L(1+\gamma)^{2}}{(\mu-\lambda)^{2}}+\frac{3L}{2\mu(\mu-\lambda)}\right)\left\|\nabla f(\mathbf{\mathbf{w}}^{t-1})\right\|^{2}\nonumber\\&+\left(\frac{L(1+\gamma)^{2}}{(\mu-\lambda)^{2}}+\frac{L}{\mu(\mu-\lambda)}+\frac{\gamma}{\mu-\lambda}\right)\mathbb{E}_{S_{t}}\left[\left\|\mathbf{g}_{t}-\nabla f(\mathbf{\mathbf{w}}^{t-1})\right\|^{2}\right],
	\end{align}}
	
	where in the last step, we used the inequality in \eqref{eq:ineq}.

Note that 
{\small\begin{align}
    \mathbb{E}_{S_t} \left[\left\|\mathbf{g}_t - \nabla f(\bw^{t-1})\right\|^2\right] = \mathbb{E}_{S_t}\left[\|\mathbf{g}_t\|^2\right] - \left\|\nabla f(\bw^{t-1})\right\|^2 \leq \mathbb{E}_k\left[ \| \nabla F_k(\bw^{t-1}) \|^2\right] - \left\|\nabla f(\bw^{t-1})\right\|^2 \leq (B^2-1) \left\|f(\bw^{t-1})\right\|^2.
\end{align}}

Plugging into \eqref{eq:th2_pr_4}, we get
{\small\begin{align}
    \mathbb{E}_{S_t}\left[f(\bw^t)\right] \leq f(\bw^{t-1}) - \rho \left\|\nabla f(\bw^{t-1})\right\|^2,
\end{align}}

where
{\small\begin{align}
    \rho = \frac{1}{\mu}-\frac{3\gamma}{2(\mu-\lambda)} - \frac{L(1+\gamma)^2}{(\mu-\lambda)^2} - \frac{3L}{2\mu(\mu-\lambda)} - \left(B^2-1\right) \left(\frac{L(1+\gamma)^2}{(\mu-\lambda)^2}+\frac{L}{\mu(\mu-\lambda)}+\frac{\gamma}{\mu-\lambda}\right).
\end{align}}
\end{proof}
	
\subsection{Proof for Theorem~\ref{th:device_specific}}\label{appen:theorem7}
\begin{proof}
	We have by the Lipschitz continuity of the gradients:
	{\small\begin{align}
	\label{eq:th4_pr_1}
	&f(\mathbf{\mathbf{w}}^{t}_{k}) \leq f(\mathbf{\mathbf{w}}^{t-1}) + \langle \nabla f(\mathbf{\mathbf{w}}^{t-1}), \mathbf{\mathbf{w}}^{t}_{k}-\mathbf{\mathbf{w}}^{t-1}\rangle + \frac{L_{k}}{2}\left\|\mathbf{\mathbf{w}}^{t}_{k}-\mathbf{\mathbf{w}}^{t-1}\right\|^{2}\nonumber\\
	&\leq f(\mathbf{\mathbf{w}}^{t-1}) + \langle \nabla f(\mathbf{\mathbf{w}}^{t-1}), \underline{\mathbf{w}}^{t}_{k}-\mathbf{\mathbf{w}}^{t-1}\rangle+ \langle \nabla f(\mathbf{\mathbf{w}}^{t-1}), \mathbf{\mathbf{w}}^{t}_{k}-\underline{\mathbf{w}}^{t}_{k}\rangle + \frac{L_{k}}{2}\left\|\mathbf{\mathbf{w}}^{t}_{k}-\mathbf{\mathbf{w}}^{t-1}\right\|^{2}
	\end{align}}
	
	By optimality conditions, we have that $\underline{\mathbf{w}}^{t}_{k}$ satisfies
	{\small\begin{align}
	\label{eq:th4_pr_2}
	\nabla F_{k}\left(\underline{\mathbf{w}}^{t}_{k}\right)+\mathbf{g}_{t} - \nabla F_{k}\left(\mathbf{\mathbf{w}}^{t-1}\right) + \mu_{k}\left(\underline{\mathbf{w}}^{t}_{k}-\mathbf{\mathbf{w}}^{t-1}\right) = 0.
	\end{align}}
	
    Similarly, we denote the local subproblem \eqref{eq:obj} as $P_t(w)$.
	We also note that, $P_t(\bw)$ is $(\mu_{k}-\lambda)$-strongly convex, 
	{\small\begin{align}
	\label{eq:th4_pr_3}
	\mu_{k}\left\|\underline{\mathbf{w}}^{t}_{k}-\mathbf{\mathbf{w}}^{t-1}\right\| \le \left\|\nabla P_t(\mathbf{\mathbf{w}}^{t-1}) \right\| = \left\|\mathbf{g}_{t}\right\|.
	\end{align}}
	
	We derive a bound for $\left\|\mathbf{\mathbf{w}}^{t}_{k}-\mathbf{\mathbf{w}}^{t-1}\right\|$ next.
	{\small\begin{align}
	\label{eq:th4_pr_3.5}
	\left\|\mathbf{\mathbf{w}}^{t}_{k}-\mathbf{\mathbf{w}}^{t-1}\right\| \le \left\|\underline{\mathbf{w}}^{t}_{k}-\mathbf{\mathbf{w}}^{t-1}\right\|+\left\|\underline{\mathbf{w}}^{t}_{k}-\mathbf{\mathbf{w}}^{t}_{k}\right\| \le (1+\gamma_{k})\left\|\underline{\mathbf{w}}^{t}_{k}-\mathbf{\mathbf{w}}^{t-1}\right\|.
	\end{align}}
	
	Using \eqref{eq:th4_pr_2}-\eqref{eq:th4_pr_3} in \eqref{eq:th4_pr_1}, we have,
	{\small\begin{align}
	\label{eq:th4_pr_4}
	&f(\mathbf{\mathbf{w}}^{t}_{k}) \leq f(\mathbf{\mathbf{w}}^{t-1}) + \langle \nabla f(\mathbf{\mathbf{w}}^{t-1}), \underline{\mathbf{w}}^{t}_{k}-\mathbf{\mathbf{w}}^{t-1}\rangle+ \langle \nabla f(\mathbf{\mathbf{w}}^{t-1}), \mathbf{\mathbf{w}}^{t}_{k}-\underline{\mathbf{w}}^{t}_{k}\rangle + \frac{L_{k}}{2}\left\|\mathbf{\mathbf{w}}^{t}_{k}-\mathbf{\mathbf{w}}^{t-1}\right\|^{2}\nonumber\\
	&\leq f(\mathbf{\mathbf{w}}^{t-1}) - \frac{1}{\mu_{k}} \langle \nabla f(\mathbf{\mathbf{w}}^{t-1}), \nabla F_{k}\left(\underline{\mathbf{w}}^{t}_{k}\right)+\mathbf{g}_{t} - \nabla F_{k}\left(\mathbf{\mathbf{w}}^{t-1}\right)\rangle + \left\|\nabla f(\mathbf{\mathbf{w}}^{t-1})\right\|\left\|\mathbf{\mathbf{w}}^{t}_{k}-\underline{\mathbf{w}}^{t}_{k}\right\| + \frac{L_{k}(1+\gamma_{k})^{2}}{2\mu_{k}^{2}}\left\|\mathbf{g}_{t}\right\|^{2}\nonumber\\
	&\leq f(\mathbf{\mathbf{w}}^{t-1})-\frac{\nabla^{\top} f(\mathbf{\mathbf{w}}^{t-1})\mathbf{g}_{t}}{\mu_{k}}+\frac{L_{k}}{\mu_{k}}\left\|\nabla f(\mathbf{\mathbf{w}}^{t-1})\right\|\left\|\underline{\mathbf{w}}^{t}_{k}-\mathbf{\mathbf{w}}^{t-1}\right\|+\gamma\left\|\nabla f(\mathbf{\mathbf{w}}^{t-1})\right\|\left\|\underline{\mathbf{w}}^{t}_{k}-\mathbf{\mathbf{w}}^{t-1}\right\|+ \frac{L_{k}(1+\gamma_{k})^{2}}{2\mu_{k}^{2}}\left\|\mathbf{g}_{t}\right\|^{2}\nonumber\\
	&\leq f(\mathbf{\mathbf{w}}^{t-1})-\frac{\nabla^{\top} f(\mathbf{\mathbf{w}}^{t-1})\mathbf{g}_{t}}{\mu_{k}}+\frac{L_{k}}{2\mu_{k}^{2}}\left(\left\|\nabla f(\mathbf{\mathbf{w}}^{t-1})\right\|^{2}+\left\|\mathbf{g}_{t}\right\|^{2}\right)\nonumber\\
	&+\frac{\gamma_{k}}{2\mu_{k}}\left(\left\|\nabla f(\mathbf{\mathbf{w}}^{t-1})\right\|^{2}+\left\|\mathbf{g}_{t}\right\|^{2}\right)+ \frac{L_{k}(1+\gamma_{k})^{2}}{2\mu_{k}^{2}}\left\|\mathbf{g}_{t}\right\|^{2}\nonumber\\
	&\Rightarrow f(\mathbf{\mathbf{w}}^{t}) \leq \frac{1}{K_t}\sum_{k=1}^{K_t}f(\mathbf{\mathbf{w}}^{t}_{k})\leq f(\mathbf{\mathbf{w}}^{t-1})-\left(\frac{1}{K_t}\sum_{k=1}^{K_t}\frac{1}{\mu_{k}}\right)\nabla^{\top} f(\mathbf{\mathbf{w}}^{t-1})\mathbf{g}_{t}+\frac{1}{K_t}\sum_{k=1}^{K_t}\frac{L_k}{2\mu_k^{2}}\left(\left\|\nabla f(\mathbf{\mathbf{w}}^{t-1})\right\|^{2}+\left\|\mathbf{g}_{t}\right\|^{2}\right)\nonumber\\
	&+\left(\frac{1}{K_t}\sum_{k=1}^{K_t}\frac{\gamma_k}{2\mu_k}\right)\left(\left\|\nabla f(\mathbf{\mathbf{w}}^{t-1})\right\|^{2}+\left\|\mathbf{g}_{t}\right\|^{2}\right)+ \left(\frac{1}{K_t}\sum_{k=1}^{K_t}\frac{L_k(1+\gamma_k)^{2}}{2\mu_k^{2}}\right)\left\|\mathbf{g}_{t}\right\|^{2}\nonumber\\
	&\Rightarrow \mathbb{E}_{S_{t}}\left[f(\mathbf{\mathbf{w}}^{t}) \right] \le  f(\mathbf{\mathbf{w}}^{t-1})-\frac{1}{K_t}\sum_{k=1}^{K_t}\left(\frac{1}{\mu_{k}}-\frac{3\gamma_k}{2\mu_k}\right)\left\|\nabla f(\mathbf{\mathbf{w}}^{t-1})\right\|^{2}+\frac{1}{K_t}\sum_{k=1}^{K_t}\left(\frac{L_{k}(1+\gamma_{k})^{2}}{\mu_{k}^{2}}+\frac{3L_{k}}{2\mu_{k}^{2}}\right)\left\|\nabla f(\mathbf{\mathbf{w}}^{t-1})\right\|^{2}\nonumber\\&+\frac{1}{K_t}\sum_{k=1}^{K_t}\left(\frac{L(1+\gamma_{k})^{2}}{\mu_{k}^{2}}+\frac{L_{k}}{\mu_{k}^{2}}+\frac{\gamma_{k}}{\mu_{k}}\right)\mathbb{E}_{S_{t}}\left[\left\|\mathbf{g}_{t}-\nabla f(\mathbf{\mathbf{w}}^{t-1})\right\|^{2}\right],
	\end{align}}
	where $|S_{t}|=K_{t}$ and in the last step, we used the inequality in \eqref{eq:ineq}.
	
	Note that 
    {\small\begin{align}
        \mathbb{E}_{S_t} \left[\left\|\mathbf{g}_t - \nabla f(\bw^{t-1})\right\|^2\right] = \mathbb{E}_{S_t}\left[\|\mathbf{g}_t\|^2\right] - \left\|\nabla f(\bw^{t-1})\right\|^2 \leq \mathbb{E}_k\left[ \| \nabla F_k(\bw^{t-1}) \|^2\right] - \left\|\nabla f(\bw^{t-1})\right\|^2 \leq (B^2-1) \left\|f(\bw^{t-1})\right\|^2.
    \end{align}}

    Plugging into \eqref{eq:th4_pr_4}, we get
    {\small\begin{align}
        \mathbb{E}_{S_t}\left[f(\bw^t)\right] \leq f(\bw^{t-1}) - \rho \left\|\nabla f(\bw^{t-1})\right\|^2,
    \end{align}}

where
	{\small\begin{align}
	    \rho=\frac{1}{K_t}\sum_{k=1}^{K_t} \left(\frac{1}{\mu_k}- \frac{3\gamma_k}{2\mu_k}-\frac{L_k\left(1+\gamma_k\right)^2}{\mu_k^2}-\frac{3L_k}{2\mu_k^2}\right) - \frac{1}{K_t}\sum_{k=1}^{K_t} \left(\frac{L\left(1+\gamma_k\right)^2}{\mu_k^2} + \frac{L_k}{\mu_k^2}+\frac{\gamma_k}{\mu_k}\right)\left(B^2-1\right).
	\end{align}}
	
	\end{proof}

\subsection{Additional Experiments}\label{appen:exp}

\begin{figure}
    \centering
    \includegraphics[width=\textwidth]{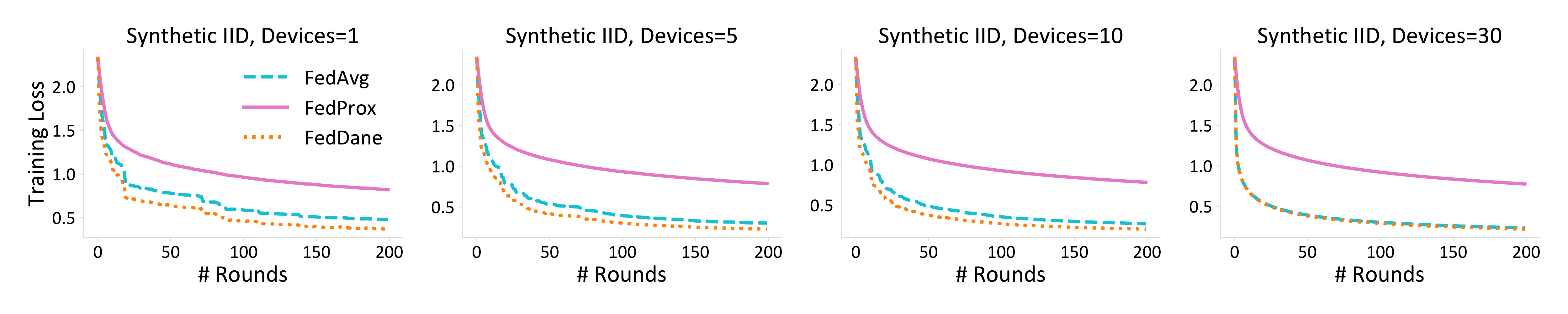}
    \includegraphics[width=\textwidth]{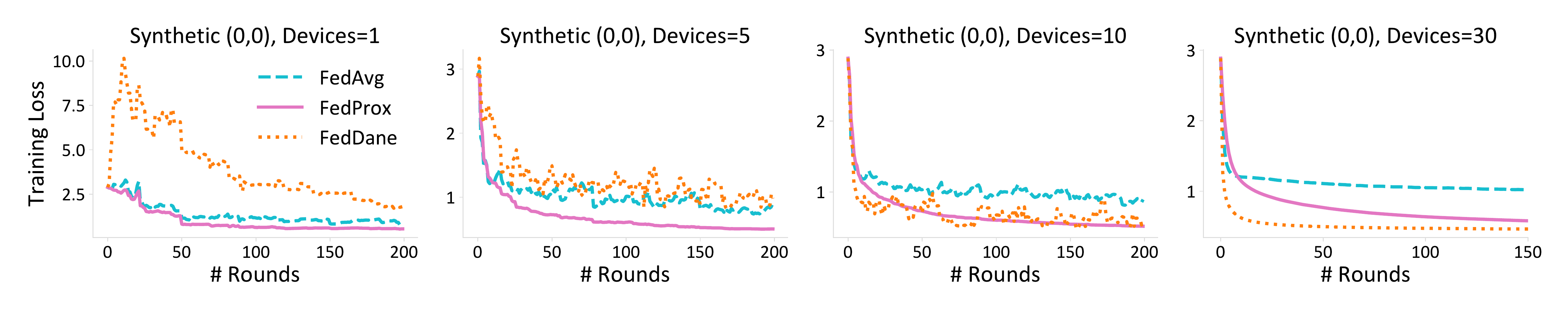}
    \includegraphics[width=\textwidth]{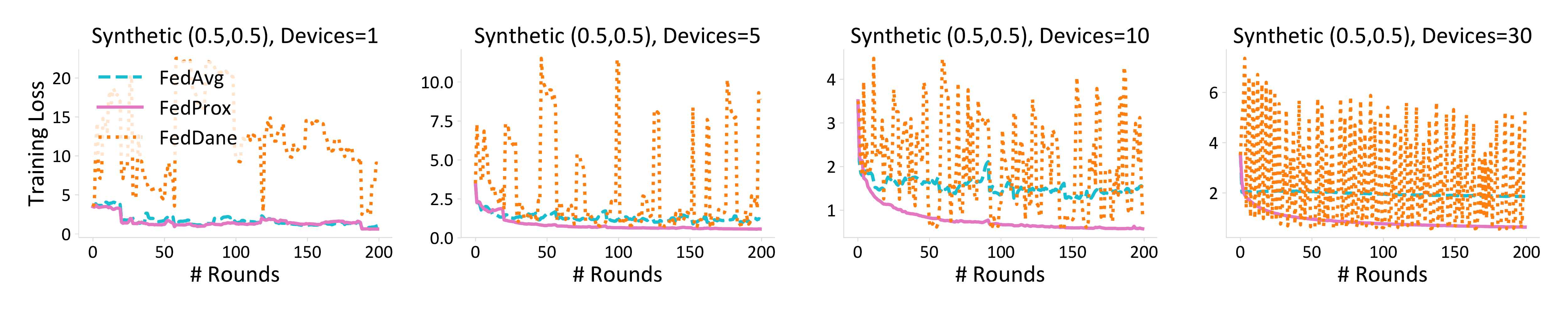}
    \caption{Effects of low device participation. For the three synthetic datasets with varying statistical heterogeneity, we randomly select 1, 5, 10, or 30 devices (out of 30) at each communication round. We set $E$ to be 20. From the top row to the bottom row, data heterogeneity is increasing. We see that (1) low device participation hurts the performance of \feddane in statistically heterogeneous settings, and (2) in highly heterogeneous environments (e.g., on the Synthetic (0.5,0.5) dataset), even full device participation does not help improve the performance of \feddane.}
    \label{fig:low}
\end{figure}

\begin{figure*}[t]
    \centering
    \includegraphics[width=\textwidth]{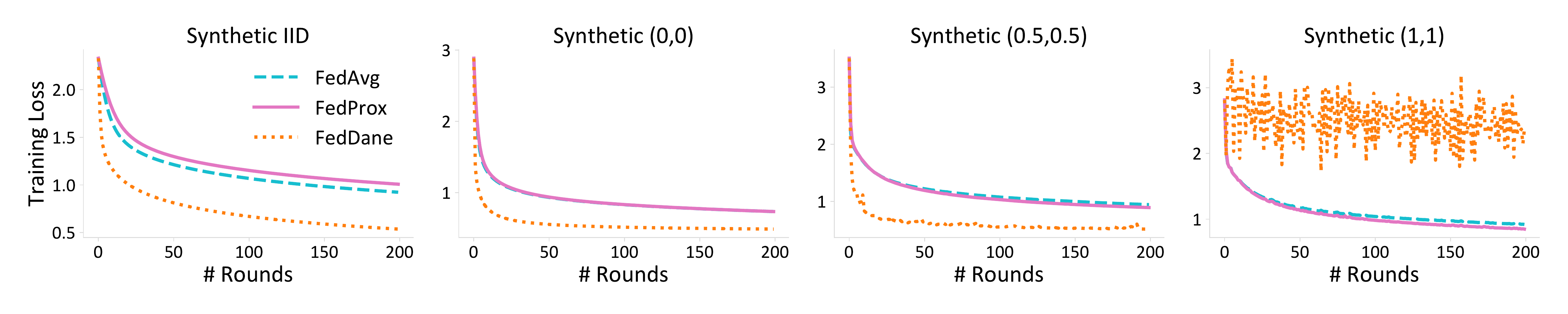}
     \includegraphics[width=0.26\textwidth]{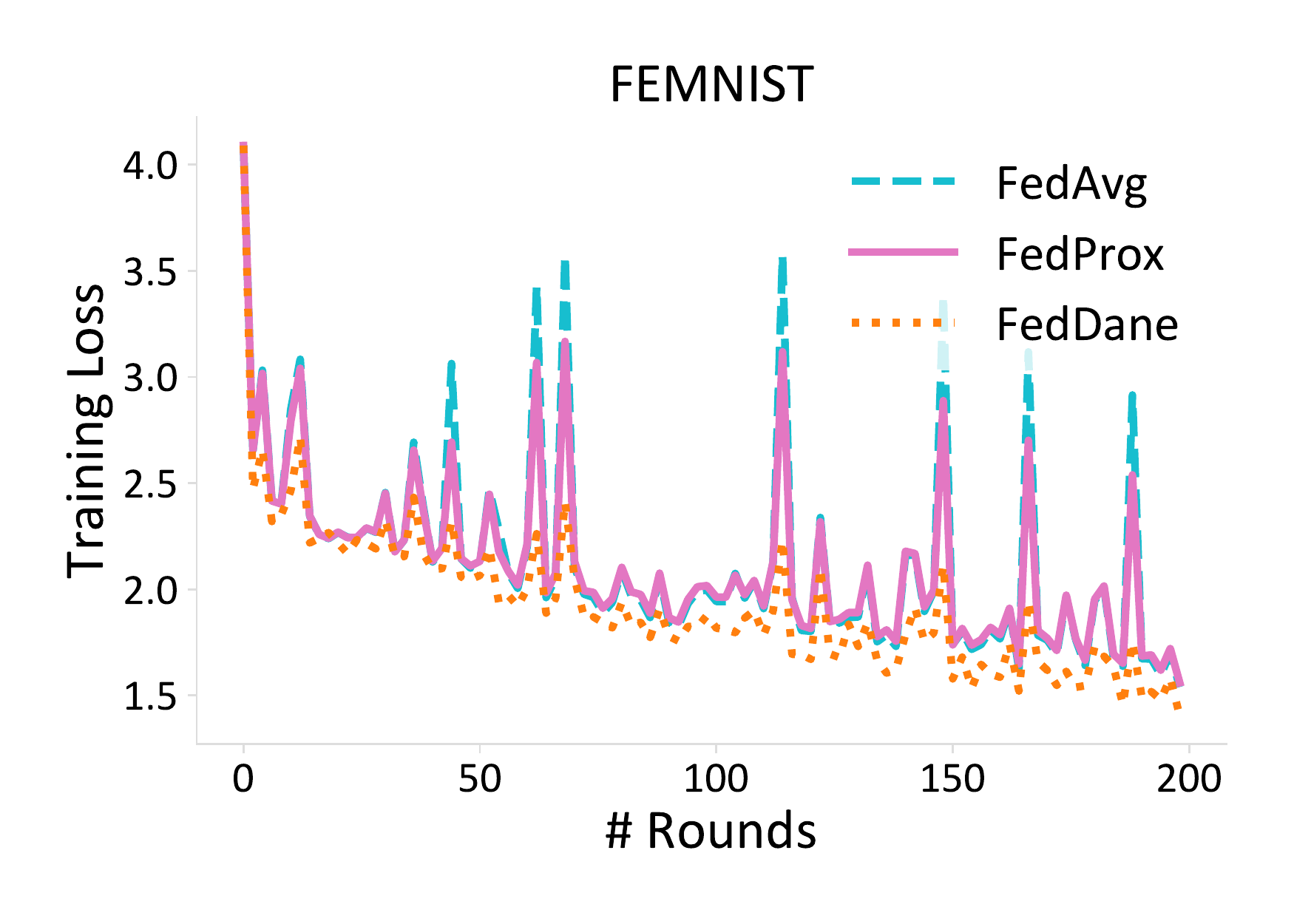}
    \includegraphics[width=0.28\textwidth]{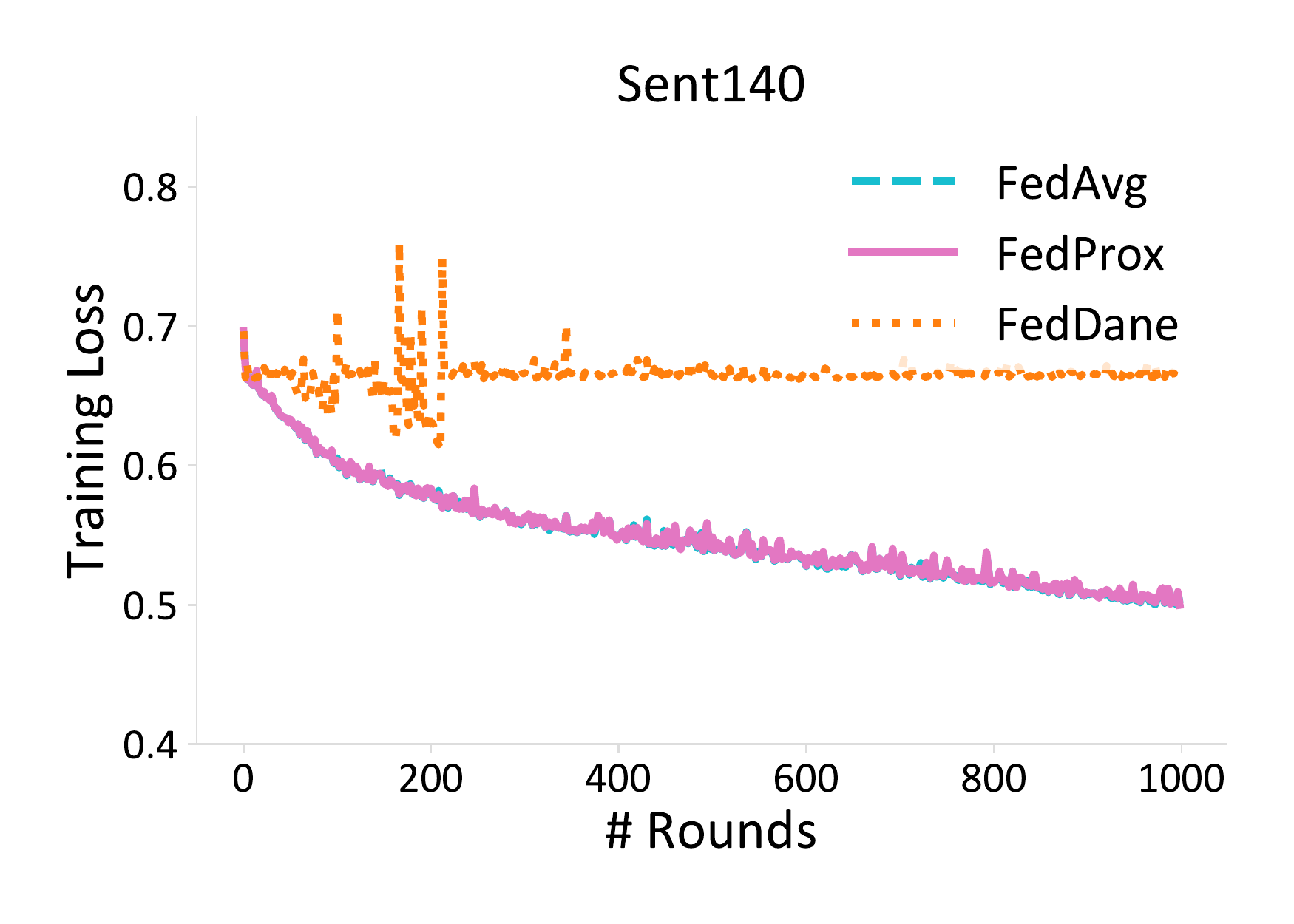}
    \includegraphics[width=0.26\textwidth]{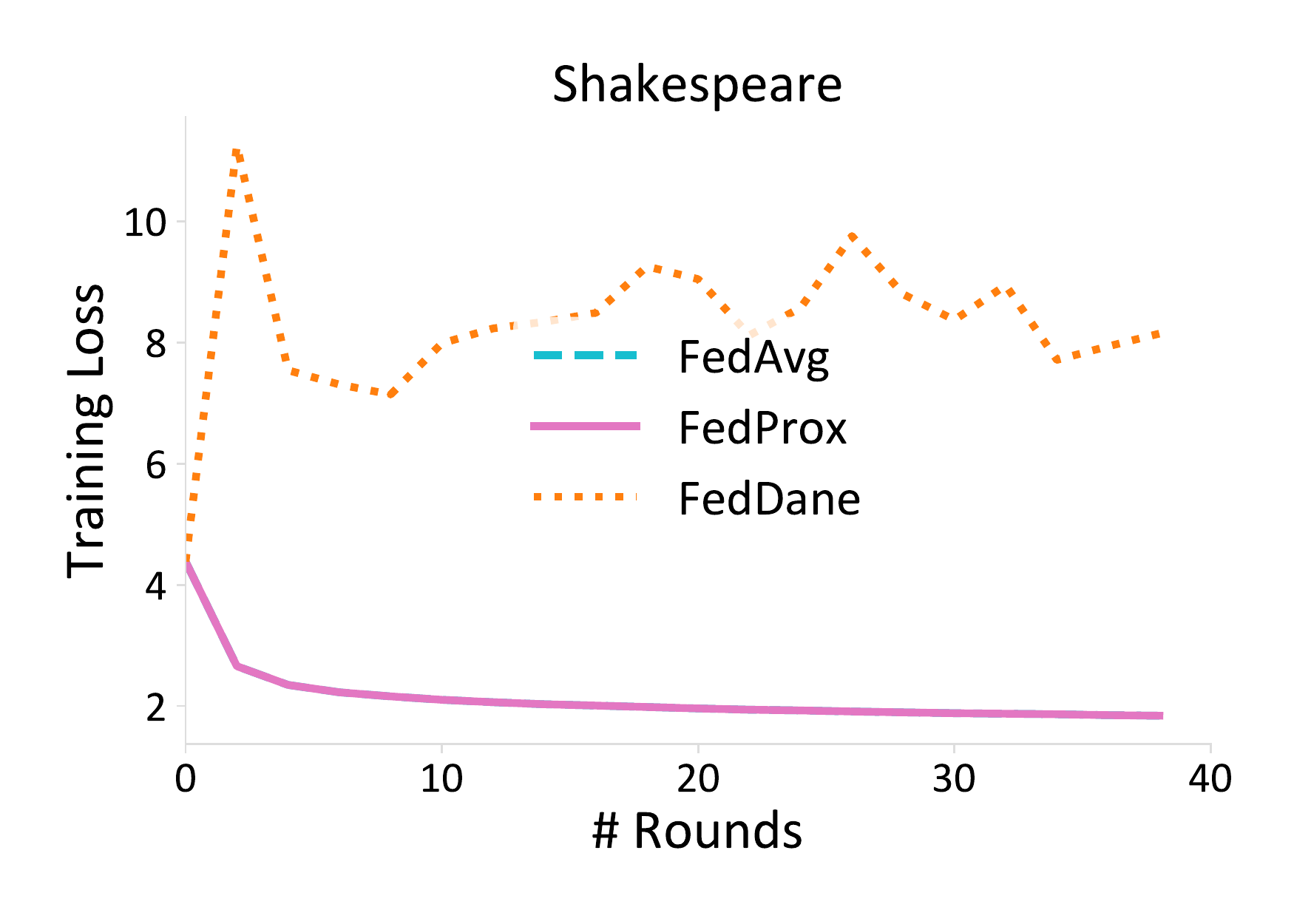}
    \caption{Convergence of \feddane compared with \fedavg and \fedprox in \emph{unrealistic settings} (nearly full device participation, small local epochs $E=1$) which favor \feddane. For synthetic datasets, we let all devices participate in learning at each iteration. For FEMNIST, Sent140, and Shakespeare, we select 50\%, 26\%, and 70\% devices respectively at each round in order to better estimate the full gradients. \feddane still performs worse than the other two methods, especially on highly heterogeneous datasets.}
    \label{fig:results2}
\end{figure*}
\end{document}